\documentclass[utf8]{frontiersSCNSmod} 

\usepackage{url,hyperref,lineno,microtype}
\usepackage[onehalfspacing]{setspace}
\usepackage{booktabs}
\usepackage[colorinlistoftodos]{todonotes}
\usepackage{amsmath}
\usepackage{amssymb}
\usepackage[mathscr]{euscript}
\usepackage{algorithm}
\usepackage{algorithmic}
\usepackage{float}
\usepackage{graphicx}
\usepackage{wrapfig}
\usepackage{siunitx}
\usepackage{hyperref}
\usepackage{booktabs}
\usepackage{color,soul}

\newcommand{\INDSTATE}[1][1]{\STATE\hspace{#1\algorithmicindent}}
 
\def\keyFont{\fontsize{8}{11}\helveticabold }
\def\firstAuthorLast{G\"unther {et~al.}} 
\def\Authors{Johannes G\"unther\,$^{1,2,*}$, Nadia M. Ady\,$^{1}$, Alex Kearney\,$^{1}$, \\ Michael R. Dawson\,$^{2, 3}$ and Patrick M. Pilarski\,$^{1,2, 3}$}
\def\Address{$^{1}$Department of Computing Science, University of Alberta, Edmonton, Alberta, Canada\\
$^{2}$Alberta Machine Intelligence Institute, Edmonton, Alberta, Canada\\
$^{3}$Department of Medicine, University of Alberta, Edmonton, Alberta, Canada}
\def\corrAuthor{Johannes G\"unther}

\def\corrEmail{gunther@ualberta.ca}

\begin{document}
\onecolumn
\firstpage{1}

\title{Examining the Use of Temporal-Difference Incremental Delta-Bar-Delta for Real-World Predictive Knowledge Architectures} 

\author[\firstAuthorLast ]{\Authors} 
\address{} 
\correspondance{} 

\extraAuth{}

\maketitle

\begin{abstract}
Predictions and predictive knowledge have seen recent success in improving not only robot control but also other applications ranging from industrial process control to rehabilitation. A property that makes these predictive approaches well suited for robotics is that they can be learned online and incrementally through interaction with the environment. However, a remaining challenge for many prediction-learning approaches is an appropriate choice of prediction-learning parameters, especially parameters that control the magnitude of a learning machine's updates to its predictions (the {\em learning rates} or \textit{step sizes}). 
Typically, these parameters are chosen based on an extensive parameter search---an approach that neither scales well nor is well suited for tasks that require changing step sizes due to nonstationarity.
To begin to address this challenge, we examine the use of online step-size adaptation using the Modular Prosthetic Limb: a sensor-rich robotic arm intended for use by persons with amputations. Our method of choice, Temporal-Difference Incremental Delta-Bar-Delta (TIDBD), learns and adapts step sizes on a feature level; importantly, TIDBD allows step-size tuning and representation learning to occur at the same time. As a first contribution, we show that TIDBD is a practical alternative for classic Temporal-Difference (TD) learning via an extensive parameter search. Both approaches perform comparably in terms of predicting future aspects of a robotic data stream, but TD only achieves comparable performance with a carefully hand-tuned learning rate, while TIDBD uses a robust meta-parameter and tunes its own learning rates.
Secondly, our results show that for this particular application TIDBD allows the system to automatically detect patterns characteristic of sensor failures common to a number of robotic applications.
As a third contribution, we investigate the sensitivity of classic TD and TIDBD with respect to the initial step-size values on our robotic data set, reaffirming the robustness of TIDBD as shown in previous papers.
Together, these results promise to improve the ability of robotic devices to learn from interactions with their environments in a robust way, providing key capabilities for autonomous agents and robots.

\tiny
 \keyFont{ \section{Keywords:} Continual Learning; Reinforcement Learning; Robot learning, Long-term autonomy, Prediction} 
\end{abstract}

\section{Predictive Knowledge For Robotics} \label{sec:intro}
Autonomous agents in the real world face many challenges when interacting with and learning from the environment around them, especially if they are deployed for extended periods of time. As the real world is non-stationary and complex, many of the challenges facing a deployed agent cannot be completely foreseen by its designers in advance. 
An agent should therefore construct its understanding of the environment using an approach that is continuous and independent, so it is empowered to adapt to its environment without human assistance. 

Predictive knowledge \citep{sutton_horde_2011,white_developing_2015} is such an approach, and allows autonomous agents to incrementally construct knowledge of the environment purely through interaction \citep{drescher_made-up_1991-1, ring_continual_1994}. In a predictive knowledge architecture, the environment is modelled as a set of forecasts about how signals of interest will behave. As an agent's actions have an effect on the environment, these forecasts about what will happen next are made with consideration to a policy of agent behaviour ({\em nexting}, as described by \cite{modayil_multi-timescale_2014}). 
In this way, these predictions can capture forward-looking aspects of the environment such as, ``If I continue moving my arm to the right, how much load do I expect my elbow servo to experience?" 
For a concrete example of predictions being used to support robot control, we consider the idea of Pavlovian control, as defined by \cite{modayil2014prediction}, wherein learned predictions about what will happen next are mapped in pre-defined or fixed ways to changes in a system's control behaviours. As a principal case study, \cite{modayil2014prediction} showed how a sensor-limited robot could use a learned prediction about an impending collision to take evasive action and reduce strain on its motors \textit{before} a collision actually occurred. Without using predictions to alter actions, a collision would need to occur before the robot would be able to take action in response to it.


Detailed demonstrations of the potential of predictive knowledge architectures in real-world domains have been offered in industrial laser welding \citep{gunther2016intelligent}, robot navigation \citep{kahn_self-supervised_2018}, animal models of partial paralysis \citep{dalrymple2019pavlovian}, and artificial limbs \citep{pilarski2013adaptive,sherstan2015collaborative, edwards_application_2016, pilarski2017communicative}. Recently, work has focused on using predictive knowledge to construct representations of state that capture aspects of the environment that cannot be described by current observations alone \citep{schlegel_general_2018}, and on accelerating the learning of predictive knowledge through the use of successor representations \citep{sherstan_accelerating_2018}.

From a computational perspective, there is strong evidence that a predictive knowledge architecture is feasible at scale. Many predictions can be simultaneously made and learned online, incrementally \citep{sutton_horde_2011}, as a system is interacting with the environment, using methods such as temporal-difference (TD) learning \citep{sutton_learning_1988} and other standard learning algorithms from the field of reinforcement learning. 
Predictive knowledge architectures have been demonstrated to scale well \citep{white_developing_2015} and to allow real-time learning \citep{modayil_multi-timescale_2014, white_developing_2015}. 

Although research to date has comprehensively established how an agent can utilize prediction learning in a broad range of environments, it is important to note that in all these previous examples, the algorithm for learning is fixed before deployment and does not itself change during learning. Specifically, the step sizes (learning rates) used by the learning algorithms in existing studies are hand-selected in advance by the experimenters through large parameter sweeps or empirical tuning. In addition to the impracticality of hand-selecting learning algorithm parameters, using a predefined and fixed step size for the lifetime of an agent might in fact significantly limit the learning capability of the agent. 

It is natural to expect that the learning rate of a long-lived agent should change over time. The process of destabilizing memories and making them prone to change is observed in mammals \citep{sinclair2018surprise} and is analogous to a temporary increase in learning rates in an autonomous agent. Such a mechanism would be especially useful in a lifelong learning setting \mbox{\citep{silver2013lifelong}}, where an agent is confronted with a changing environment and parameters cannot be optimized in advance.
Following this idea, recent research has investigated approaches capable of online step-size adaptation \citep{mahmood2012tuning,sutton1992adapting}, wherein a learning agent is able to self-tune the size of the learning steps it takes in response to the errors observed during its own learning process. However, the aforementioned step-size adaptation methods still use a single step size for all inputs and therefore treat all inputs to a learning agent equally. Not surprisingly, the reliability and variability of different inputs can play a large role in an agent's ability to learn about future outcomes---inputs are not all created equal in terms of their utility for a learning agent. The use of a single scalar step size therefore limits an agent's ability to adapt to and learn more about interesting inputs and to learn less about uninteresting or noisy inputs.

There are several learning rate adaptation methods that modify each individual step-size. AdaGrad, RMSProp, and AMSGrad are methods for deep learning. Temporal-difference learning in and of itself has no relation to neural networks. We can view a neural network as a function approximator for a TD learning method; however, this is one of many function approximators that could be chosen. For instance, the methods used in \citet{jacobsen2019meta} use a binary recoding of features, similar to the methods in this paper.
Appealing to intuition for a moment, when we consider the strengths of RMSProp, ADAM, and others, it’s predominantly useful in helping with difficult optimization surfaces for stochastic gradient descent. In part, what makes RMSProp and ADAM so successful is their ability to counteract the difficulties which arise from vanishing and exploding gradients in very deep neural networks. Understanding that TD learning is not stochastic gradient descent, and has no direct relation to deep artificial neural networks, we can then ask whether such problems apply to linear TD problems, such as the ones explored in our paper. The problems of applying these optimizers to TD learning is demonstrated in part in the empirical comparisons done by \citet{kearney_tidbd_2019}, who provided a comparison of RMSPROP and TIDBD in their Figure 11; their analysis showed RMSPROP performed worse than TIDBD on their TD-learning task. RMSProp produced empirical return errors so high, that it could not even be plotted alongside the TD-specific step-size adaptation methods.

By implementing an individual step size for each input to a learning agent, it is possible for an agent to treat different inputs differently during learning. One extension of scalar step-size adaptation methods to a non-scalar form is Temporal-Difference Incremental Delta-Bar-Delta (TIDBD) \citep{kearney_tidbd_2019}. In their introduction of TIDBD, \cite{kearney_tidbd_2019} investigated adaptation of vector step sizes on a feature level, comparing how TIDBD adapts the step sizes for noisy features versus prediction-relevant features.
In this work, we translate TIDBD to a more realistic setting. 
Rather than investigating deteriorating features, we investigate deteriorating sensors; we consider the case where a set of sensors freezes or becomes noise, preventing perception of a useful signal. Such a situation rarely translates cleanly to a simple set of unrelated feature noise in the feature representation. \cite{kearney_tidbd_2019} found in their experiments that TIDBD could outperform TD methods that lack step-size adaptation. As an extension of the work done by \cite{kearney_tidbd_2019}, we compare TIDBD against TD methods on a robot data stream with far more parallel signals than any prior test domains. In addition, we consider the viability of TIDBD on this complex data in terms of computation and memory.

A meta-learning method that can perform comparably to, or outperform, classic TD, yet avoid the need for time- and labour-intensive parameter tuning, is one main component for making predictive architectures practical in real-world applications. Although meta-learning methods promise to adapt parameters without human intervention, they themselves introduce new meta-parameters. Fortunately, the algorithmic performance is more robust with respect to these meta-parameters, as shown by \mbox{\cite{mahmood2012tuning} and \cite{kearney_tidbd_2019}.} This robustness allows these meta-parameter to be set with default values, rendering a parameter search unnecessary.

As main contributions of this work, we provide deeper understanding and intuition about the effect that using TIDBD will have on prediction-learning tasks involving complex, real-world data. We furthermore investigate the robustness of TIDBD on our robotic data set in order to reaffirm previous experiments. In what follows, we demonstrate how TIDBD adapts the step sizes in TD learning when confronted with a non-stationary environment. By examining the operation of TIDBD in comparison to classic TD and its ability to perform feature selection in relation to specific signals in the robotic arm, this work carves out insight that will help others design persistent agents capable of long-term autonomous operation and learning.
\section{Prediction-Learning Methods}
Key to the construction of predictive knowledge systems is the way predictions are specified. One proposal is to express world knowledge as a collection of General Value Functions (GVFs) \citep{sutton_horde_2011}.
Interaction with the world is described sequentially, where at each time step $t$, an agent takes an action $A_t \in \mathscr{A}$, which causes a transition from $S_t$ to $S_{t+1} \in \mathscr{S}$ assumed to be specified by a Markov Decision Process. The agent's choice of action, $A_t$, is determined by a probability function $\pi: \mathscr{S} \times \mathscr{A} \longrightarrow [0,1]$, known as a \textit{policy}. We model our world by forming predictive questions about our sensations, which we phrase as GVFs---predictions about a signal of interest, or \textit{cumulant}, $C$, from the environment over some time-scale or horizon defined by $\gamma \geq 0$, and some behaviour policy $\pi$. The discounted future sum of the cumulant is known as the \textit{return}, $G_t = \sum^{\infty}_{k=0}\gamma^k C_{t+k+1}$. A GVF, $V$, is the expected return of this cumulant: $V(s;\pi,\gamma,C) = \mathbb{E}[G_t | S_t = s]$, which can be estimated using incremental online learning methods, such as TD learning \citep{sutton2018reinforcement}. A collection of GVF learners is called a \textit{Horde} \citep{sutton_horde_2011}.

In complex domains, such as the robotics domain we explore in this paper, the state space can be large or infinite: we must use function approximation to represent GVFs. The most common form of function approximation that has been used in fundamental work with GVFs is linear function approximation \mbox{\citep{modayil_multi-timescale_2014, white_developing_2015, sutton_horde_2011}}. 
Furthermore, there are multiple demonstrations showing the usability of linear function approximation in combination with GVFs in different applications, such as laser welding \mbox{\citep{gunther2018machine}} and prosthetic limbs \mbox{\citep{edwards_application_2016}}.
For these reasons, we use a linear function approximation approach in this work.
%
To construct our features, we use selective Kanerva coding, a method shown by \cite{travnik2018reinforcement} to be less sensitive to the curse of dimensionality than tile coding, yet still offering linear complexity. As our state space has $108$ dimensions, this is an important advantage. Selective Kanerva coding represents the state space with a number of \textit{prototypes}. These prototypes are points in the state space which are typically distributed throughout the state space based on a uniform random distribution. 
In the vanilla Kanerva coding algorithm, a state would be represented by all prototypes within a defined radius around a state. This leads to a variable number of active prototypes, resulting in high-variance updates for the value function. However, in \textit{selective} Kanerva coding, a state (point in this space) is represented by a constant number of the closest prototypes, providing a binary feature vector indexed by the prototypes. The constant number of active prototypes leads to less variance in the magnitude of the update.

When performing TD learning with linear function approximation, we estimate the value $V(s)$ as the dot product of a weight vector $\textbf{w}$ with a feature vector $\textbf{x}(s)$, constructed through selective Kanerva coding to act as the state representation of $s$. We improve our estimate $\textbf{w}^\top\textbf{x}(s)$ of $V(s)$ through gradient descent in the direction of a TD error $\delta_t = C_{t+1} + \gamma V(S_{t+1}) - V(S_t)$ (Algorithm \ref{TIDBD_alg}, Line 3). The weights, $\textbf{w}$, for each GVF learner are updated on each time step. 

\subsection{AutoStep TIDBD}
\label{sec:TIDBD}

When learning GVFs, two perennial challenges are setting parameters and choosing adequate representations. 
Oftentimes a representation is chosen as a fixed set of features, and this set of features and appropriate parameter settings are selected based on the results of extensive sweeps over large datasets.  
In the online lifelong continual learning setting, such a sweep is an impossible task. When we expect learning to occur over the life of the agent in non-stationary environments, we cannot ensure that a setting appropriate for only a subsample of experience will be appropriate in all situations.

\begin{algorithm}[t!]
\caption{TD($\lambda$) with AutoStep TIDBD($\lambda$)}
\begin{algorithmic}[1]
\STATE Initialize vectors $\textbf{h} = {0}^{n}$ (which will act as a decaying trace of recent weight updates), $\textbf{z} = {0}^{n}$ (which will act as a decaying trace of recently active features), and both $\textbf{w} \in \mathbb{R}^{n}$ and $\boldsymbol\beta \in \mathbb{R}^{n}$ as desired, and set $\alpha_i = e^{\beta_i}$ for each element $i = 1, 2, \ldots, n $; initialize scalars $\theta>0$ and $\tau > 0$ as appropriate (see Section \mbox{\ref{sec:TIDBD}} for suggested settings); observe state $S_{t}$
\STATE Repeat for each observation $S_{t+1}$ and cumulant $C$: 
  \INDSTATE[1] Construct feature representation with $x_i(S_t)$ as $i$th element of $\textbf{x}(S_t)$ for each element $i = 1, \ldots, n$
  \INDSTATE[1] $\delta \gets C + \gamma \textbf{w}^\top \textbf{x}(S_{t+1}) - \textbf{w}^\top \textbf{x}(S_{t})$
  \INDSTATE[1] For element $i = 1, 2, \ldots, n $: 
    \INDSTATE[2] $\xi_i \gets \max( $\\ 
    \INDSTATE[4] $|\delta [\gamma x_i(S_{t+1}) - x_i(S_{t})] h_i|,$ \\
    \INDSTATE[4] $\xi_i - \frac{1}{\tau} \alpha_i [\gamma x_i(S_{t+1}) - x_i(S_{t})]  z_i[|\delta x_i(S_{t}) h_i| - \xi_i]$ \\ 
    \INDSTATE[3] $) $
  	\INDSTATE[2] $\beta_i \gets \beta_i - \theta \frac{1}{\xi_i}\delta [\gamma x_i(S_{t+1})) - x_i(S_{t})] h_i$
    \INDSTATE[1] $M \gets \max(-e^{\beta} [\gamma \textbf{x}(S_{t+1}) - \textbf{x}(S_{t})]^\top \textbf{z}$, 1)
    \INDSTATE[1] For element $i = 1, 2, \ldots, n $: 
  
    \INDSTATE[2] $\beta_i \gets \beta_i - \log(M)$
    \INDSTATE[2] $\alpha_i \gets e^{\beta_i}$
  	\INDSTATE[2] $z_i \gets z_i  \gamma \lambda + x_i(S_{t})$
  	\INDSTATE[2] $w_i \gets w_i + \alpha_i \delta z_i$   
  	\INDSTATE[2] $h_i \gets h_i \max(1 + \alpha_i z_i [\gamma x_i(S_{t+1}) - x_i(S_{t})]  , 0) + \alpha_i\delta z_i$ 
  \INDSTATE[1] $s \gets S_{t+1}$
\end{algorithmic}
\label{TIDBD_alg}
\end{algorithm}

This difficulty of setting the parameters for a single GVF is further compounded when we expect many predictions to be learned, as is the case in predictive knowledge architectures. In particular, no single step size (also known as \textit{learning rate}) will be appropriate for every prediction in a Horde, and no single feature representation will be appropriate for every prediction in a Horde. As we cannot identify the best representations and parameters beforehand, it would be ideal to be able to adapt them through experience. To this end, we explore TIDBD: a meta-learning method for temporal-difference learning that adjusts the step-size parameter on a per-feature basis. By adapting step sizes through stochastic meta-descent, TIDBD not only provides a method of tuning parameters, but also a form of representation learning.

When hand-tuning each GVF, a single appropriate step size $\alpha > 0$ is typically chosen, resulting in the use of the same step size for every feature. TIDBD, however, adjusts a vector of many step sizes---one step size for each feature. For a binary feature vector $\textbf{x}(s) \in \mathbb{R}^n$, for example, there would be a corresponding vector of learned step sizes $\boldsymbol\alpha \in \mathbb{R}^n$, where $n$ is the number of features. At its core, TIDBD is performing stochastic meta-descent to modify the step-size parameters to minimize the squared TD error. This meta-descent results in step-size adaptation based on each feature's relevance; features which are highly correlated to the prediction problem should be given large step sizes, while irrelevant features should contribute less to updates and be given smaller step sizes.



TIDBD can be thought of as having two components 1) IDBD-based meta-learning and 2) AutoStep-based normalization. IDBD-based meta-learning provides the updates of each step size $\alpha_i$ by learning the meta weights $\boldsymbol\beta$ through meta-descent (see update in Algorithm \mbox{\ref{TIDBD_alg}}, line 10). Intuitively, this means that the more correlated recent weight updates are for a given feature $x_i$, the more effective it is to make a single large update, and thus the greater the value of the step size $\alpha_i$. This correlation is tracked using the decaying trace of recent weight updates, $\textbf{h}$ (updated in line 17). For more intuition on how IDBD abstracts the problem of tuning, see Section 2 (IDBD) of \mbox{\cite{kearney_tidbd_2019}}.

AutoStep-based normalization $\xi$ is a an additional factor for the meta weight update (see line 10). By adding AutoStep-based normalization, we 1) ensure that the effective step size is not so large that we overshoot on a given update for a particular example (line 11), and 2) maintain a running average of recent weight updates to ensure that the step sizes do not diverge if many successive weight updates are correlated (lines 6-9). The variable $\xi$ acts as a weight-update normalizer: we take the maximum value between the absolute value of the most recent weight update 
(line 7), and a decaying trace of recent updates (line 8), where $\tau$ is a parameter that determines the rate of decay. We take a decaying trace of $\xi$, as it ensures the normalizer can recover gracefully from outliers and extreme values.

 \begin{figure}[!t]
\centering
\includegraphics[width = 0.55\columnwidth]{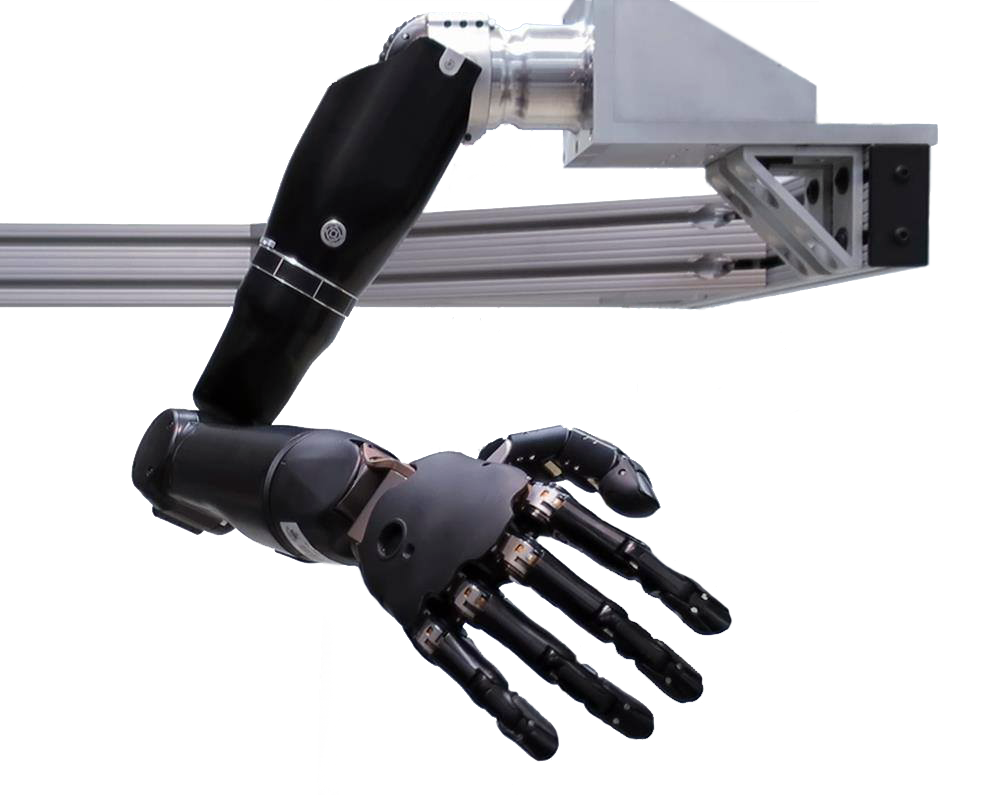}
\caption{The Modular Prosthetic Limb (MPL), a robot arm with many degrees of freedom and sensors used for the experiments in this work.}
\label{fig:mpl}
\end{figure}
 
While a common criticism of meta-learning is the introduction of additional meta parameters, TIDBD is insensitive to both of its meta parameters: the meta step size, $\theta$, and the rate-of-decay parameter, $\tau$. \mbox{\cite{kearney_tidbd_2019}} suggest $\theta = 10^{-2}$ to be a good choice over a variety of different prediction problems \mbox{\citep{kearney_tidbd_2019}}. To confirm this suggestion, we performed a robustness study, described in Section \mbox{\ref{sec:parameter_sensitivity}}. As shown by \mbox{\cite{mahmood2012tuning}}, the rate-of-decay parameter $\tau$ does not have a significant influence on performance. We therefore set it to the suggested value of $\tau=10^4$.

\section{Experimental Setup} \label{sec:experiment}
\begin{figure}[!t]
\centering
\includegraphics[width = 0.8\columnwidth]{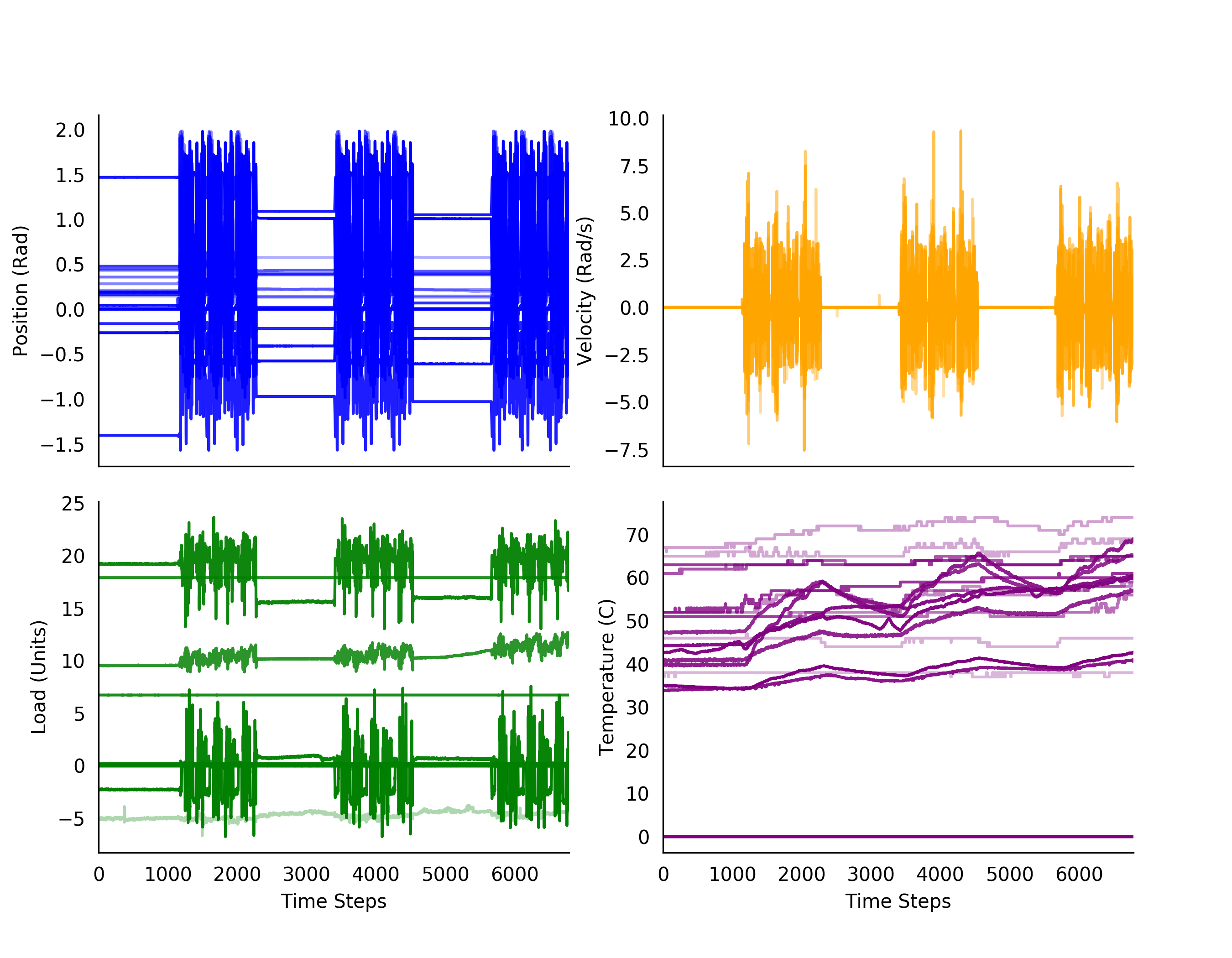}
\caption{Decoded percept data from the robot over the 30 minutes of the experiment. The periods of the arm resting and the periods of the arm moving are clearly distinguishable for the position, velocity and load sensors. The values of the temperature sensors increase over the experiment, with additional increases during the periods of movement.}
\label{fig:datastream}
\end{figure}

We gathered the data for our experiments from the Modular Prosthetic Limb (MPL v3) \citep{bridges_control_2011}---a state-of-the-art bionic limb, seen in Figure \ref{fig:mpl}, which is capable of human-like movements. The MPL includes $26$ articulated joints in its shoulder, elbow, wrist, and hand. It provides $17$ degrees of freedom. Each motor has sensors for load, position, temperature, and current; each fingertip is outfitted with a $3$-axis accelerometer and with $14$ pressure pad sensor arrays. Together, these provide a data stream of $108$ real-valued sensor readings that is shown in Figure \ref{fig:datastream}. 

Experiments by \citet{pilarski2013adaptive} suggest that real-time prediction learning can make the control of artificial limbs more intuitive for the user. In particular, anticipation and adaptation are highly important given the world and tasks encountered by a prosthetic limb are continuously changing. Therefore, the arm is an interesting showpiece as an autonomous learner \citep{pilarski2017communicative}.

In this work, the sensor readings acted as both the cumulants for predictions that the system was tasked with learning and as the state information provided to the system. Since our data included $108$ sensor readings, the architecture of the system was a Horde of $108$ predictions (general value functions), with each prediction estimating the expected discounted return for a different sensor reading. For each prediction, we used a discount rate of $\gamma = 0.9$. 
This discount rate can be thought of as resulting in a 10-step time scale, because it refers to, in expectation, the sum of the cumulant over 10 time steps.
Succeeding time steps were $0.265$ seconds apart, on average, so a $10$-step time scale refers to $2.65$ seconds. This value is potentially well suited to capturing some comparatively slow movements, e.g., the elbow extension or flexion, but might result in averaging over very fast movements, e.g., a fast grasping movement.
As the computations for the predictions and ideal return were undertaken offline---though the computations for the predictions used online methods---the computation time did not affect the length of the time step.

As state information to update each prediction, the learning system has access to all $108$ sensor readings at the current time step. These sensor readings are normalized and fed into the selective Kanerva coder to obtain feature vector $\textbf{x}(s)$. For intuition on the Kanerva coder, one can think of a $108$-dimensional cube, with each side being the $\left[ 0, 1 \right)$ interval. There are a constant number, $n$, of fixed prototypes (points) within the cube. Each normalized value fed into the Kanerva coder is a point in the cube, and can be represented by the $\eta$ nearest prototypes. We call this set of nearest prototypes \textit{active}.

Note that both classic TD and TIDBD are online methods for estimating general value functions, meaning they predict the expected return based on only information observed up to the time step they are estimating the value for. They use previous estimates, the current cumulant, and the current feature vector to make an educated guess about the future of the cumulant signal: the discounted return. A perfect estimate of the return could only be made with cumulant readings that have yet to be made.

We measured the performance of TIDBD through a comparison with classic TD. In particular, we considered the root mean squared error (RMSE), which is essentially a measurement of the difference between the true return and the value (the expected value of the return). We computed the RMSE over all predictions for a single time step, $t$, as follows. 
\begin{equation}
    \textit{RMSE}_t = \sqrt{\frac{\sum_{i=1}^{108}\left( \frac{G_t^{(i)} - \textbf{x}(s_t)^\top\textbf{w}_t^{(i)}}{\left\lvert G_t^{(i)}\right\rvert} \right)^2}{108}}
\end{equation}
The superscript $(i)$ denotes association with the $i$th prediction of 108. Normalization (division of the return and value estimate by ${\scriptstyle \left\lvert G_t^{(i)}\right\rvert}$) was done to make the RMSE meaningful, as the returns (and associated predictions) for different sensors were on different scales. 
Note that it would be unlikely for the RMSE to reach zero. While the return was computed taking all sensor readings for the full experiment into account, both classic TD and TIDBD used only sensor readings up to the present time step. These observations do not provide enough information to perfectly predict the future. 


For a meaningful comparison of classic TD with TIDBD, the parameters needed to be carefully tuned. The best parameters were chosen based on minimizing the RMSE over all predictions, summed over all time steps of the experiment. We therefore performed parameter sweeps for the number $n$ of prototypes in the Kanerva coder, the ratio $\eta$ of active prototypes to the total number of prototypes, and for the scalar step sizes $\alpha$ for each prediction for classic TD. The candidates for each parameter are shown in Table \ref{tab:parameter_settings}. The candidates for $n$, the number of prototypes, were chosen based on the recommendations provided by \cite{travnik2017representing}. 
We used a full factorial experimental setup, resulting in $264$ different parameter settings for the experiments with a fixed step size, those for classic TD. Because the TIDBD experiments did not require a sweep over potential step sizes $\alpha$, there were only $24$ different parameter settings for the TIDBD experiments (accounting for candidates for $\eta$ and $n$). In total, we conducted $288$ different experiments for our comparison of TD learning and TIDBD. 
\begin{table}[ht]
    \centering
    \begin{tabular}{c|c|l}
        Parameter   & Count   & Candidates \\ \hline \hline
        $n$         & 4           & 10000, 20000, 30000, 40000   \\
        $\eta$      & 6           & 0.001, 0.002, 0.004, 0.008, 0.016, 0.032   \\
        $\alpha$    & 11          & $\frac{0.001}{n \cdot \eta}$, $\frac{0.002}{n \cdot \eta}$, $\frac{0.004}{n \cdot \eta}$, $\frac{0.008}{n \cdot \eta}$, $\frac{0.016}{n \cdot \eta}$, $\frac{0.032}{n \cdot \eta}$, $\frac{0.064}{n \cdot \eta}$, $\frac{0.128}{n \cdot \eta}$, $\frac{0.256}{n \cdot \eta}$, $\frac{0.512}{n \cdot \eta}$, $\frac{1.024}{n \cdot \eta}$   \\
    \end{tabular}
    \caption{Parameter candidates tested in full factorial design.}
    \label{tab:parameter_settings}
\end{table}


To set the parameters for classic TD, the experiment was first run with a fixed step size shared by all GVFs in the Horde. In these experiments, using $n = 30000$ and $\eta = 0.032$ yielded the lowest RMSE in comparison to other Kanerva coder parameter choices, regardless of the choice of step size. In a second step, the RMSE for each GVF was calculated for each step size candidate, so the best step size for each GVF could be chosen independently. The best step sizes ranged from $\frac{0.001}{n\cdot\eta}$ to $\frac{0.256}{n\cdot\eta}$, where the product~$n\cdot\eta$ is the number of active features.

The parameters that yielded the best performance in terms of RMSE for classic TD also performed best in the parameter sweep for TIDBD.
For each feature, the step size was initialized to $0.00104$, which corresponds to an initial value of $\frac{1}{n\cdot\eta}$.
After the best parameters were established for classic TD and TIDBD, 30 independent trials were performed for each. 
%

%

%
We programmed the robotic arm with a repeating series of periods of rest and periods of motion. The experiments started with the arm holding its position for five minutes. This period of rest was followed by five minutes of the arm repeating a complex pattern of movement that was programmed using a data glove. The movement pattern included motion of all joints and involved movements that humans with intact arms take for granted, like grasping or flexing one finger after another. For a better understanding, the exact movement pattern can be found online at
\href{https://blinclab.ca/mpl_teleop_video/}{\texttt{https://blinclab.ca/mpl\_teleop\_video/}}. The movement pattern was $100$ seconds long, so was repeated three times during the five-minute period of movement. The periods of rest and movement alternated three times, totalling $30$ minutes.

During the rest period, each position, velocity and load sensor would be expected to report a constant signal, up to machine precision. Such sensor values should be easy to learn. During the movement pattern, on the other hand, the robot is in contact with human intention, so the predictions become far more difficult to make.
The full series of periods of rest and movement provided an interesting test case, approximating intermittent stationarity and non-stationarity, to investigate the effect of TIDBD on GVF predictions. 

Beyond our investigation of TIDBD with all sensors fully functioning, we also investigated how TIDBD reacts when confronted with two commonly occurring sensor failures: 1) sensors being stuck and 2) sensors being broken. A stuck sensor typically outputs a constant signal with a small amount of sensor noise \citep{li2012fault}, while a broken sensor typically outputs Gaussian noise with a high variance \citep{ni2009sensor}. In both experiments, the signals from all four sensors in the elbow were replaced: in the first, with Gaussian noise of $\mathcal{N}(1,0.5)$ for the stuck sensors, and with Gaussian noise of $\mathcal{N}(0,10)$ for the broken sensors.

Lastly, to investigate the robustness of TIDBD with respect to the initial step sizes $\alpha$ and its meta step size $\theta$, we performed sweeps over the data set with different values for the initial step sizes from Table\mbox{ \ref{tab:parameter_settings}} within the same range as for TD and with different meta step sizes in the range of $\theta = \{0.005, 0.01, 0.02, 0.04, 0.08, 0.16\}$.
\section{Results and Discussion}
The experiments were designed to investigate the effect that TIDBD has on predictions about the signals provided by a sensor-rich robotic arm. As a baseline, classic TD with an extensive parameter search was implemented. Three different scenarios were introduced: the predictions for different patterns of movement and rest, the predictions for the same patterns when the four elbow sensors are stuck and report a slightly noisy constant signal, and the predictions for the patterns when the four elbow sensors are broken and only report noise.

\subsection{Comparison of Classic TD and TIDBD}
\label{sec:comparison}

\begin{figure*}[!t]
    \centering
    \includegraphics[width = 1\textwidth]{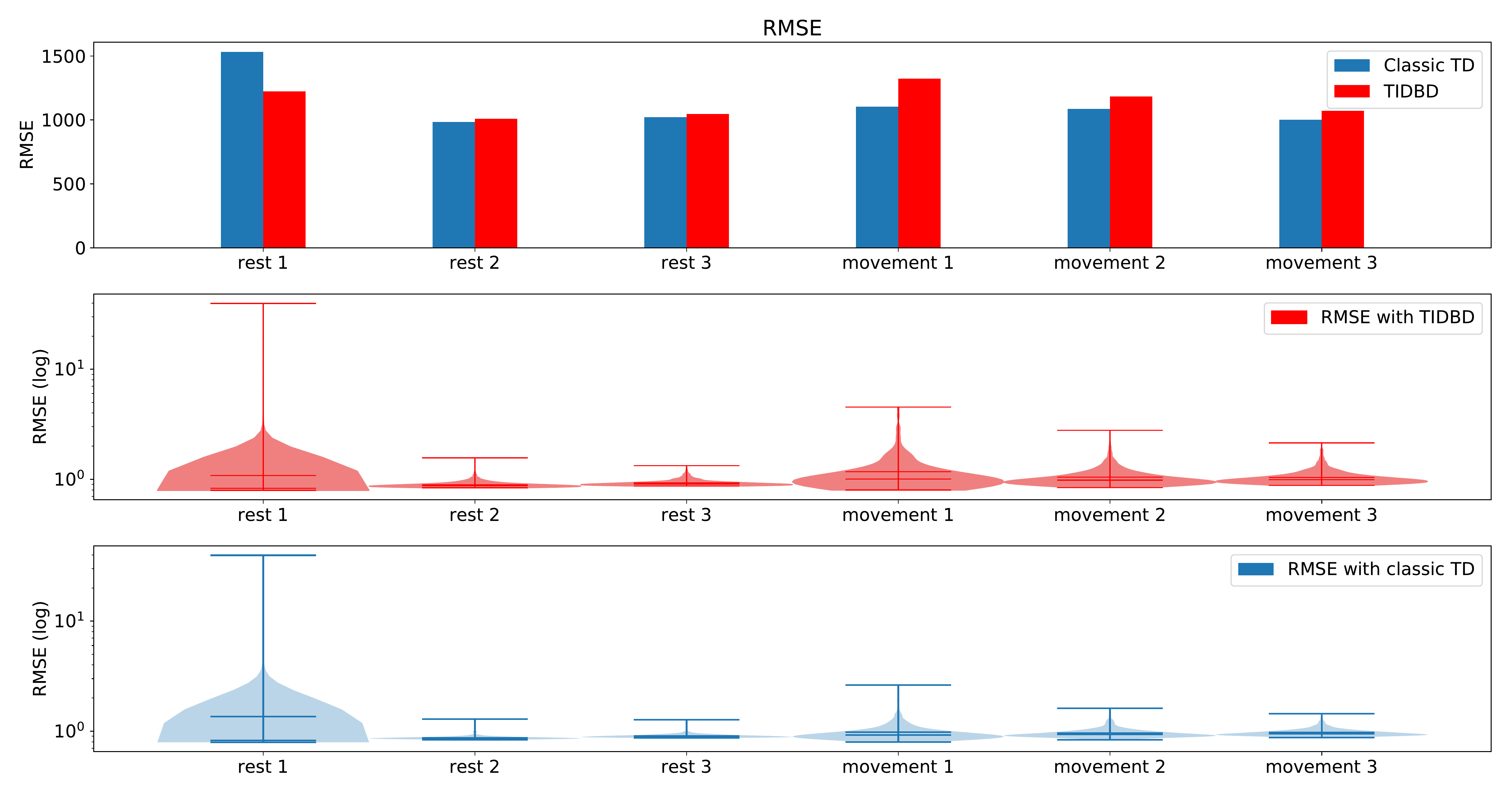}
    \caption{RMSE and violin plots for the experiment in Section \mbox{\ref{sec:comparison}}. The top pane shows the RMSE for both classic TD and TIDBD for each of the different time periods. The middle and bottom panes show violin plots for the RMSE, for TIDBD and classic TD respectively. All results are the average over 30 independent runs.}
    \label{fig:rmse_bar}
\end{figure*}
\begin{table}[ht]
    \centering
    \begin{tabular}{c|c|c}
        GVF setting     & Fixed step size   & TIDBD     \\
        \hline
        \hline
        Rest 1          & 1531.84           & 1222.90   \\
        Rest 2          & 983.72            & 1009.14   \\
        Rest 3          & 1023,00           & 1046.66   \\
        \hline
        Movement 1      & 1105.32           & 1323.94   \\
        Movement 2      & 1085.81           & 1184.09   \\
        Movement 3      & 1003.21           & 1072.13   \\

    \end{tabular}
    \caption{Average RMSE over 30 independent runs}
    \label{tab:average_rmse}
\end{table}
We first consider the root mean squared error (RMSE, as defined in Section \ref{sec:experiment}) for both classic TD and TIDBD in our initial experiment, where all sensors are fully functional. The top pane of Figure \ref{fig:rmse_bar} shows the RMSE for each period of rest and movement. It can be seen that the highest error for both classic TD and TIDBD occurs during the first period (Rest 1). This can easily be explained by all GVFs being initialized without any knowledge about the sensor readings---the RMSE for the first time steps will therefore be high. 
These high errors can be well seen in the violin plots, in the middle and bottom panes, which show the distribution over the errors the predictions made, the extrema, and the medians in the subplots for both classic TD and TIDBD. The maxima for TD and TIDBD were considerably higher in this period than for any other part of the experiment and the error distribution was much broader, as indicated by the colored area in the violin plots in Figure \ref{fig:rmse_bar}. Unsurprisingly, TIDBD exhibited a higher RMSE than classic TD at the beginning of the experiment, as its step sizes were initialized more aggressively and were not tuned to the predictive task.

The error for the second rest period was already considerably lower. Perhaps unintuitively, the error for the third rest period increased again. This can be explained by the sensor data from Figure \ref{fig:datastream}. One of the load sensors started to drift in the third rest period. As this pattern had not been seen in any of the rest periods before, the RMSE peaked again---the pattern of a drifting sensor had not been learned, yet. For the periods of movement, a steady decrease in RMSE was observed. 

On average, TIDBD had a slightly higher RMSE for the $30$-minute experiment. The exact errors for each period of rest and movement can be found in Table \ref{tab:average_rmse}. It is important to recognize that our parameter sweep over step sizes provided an advantage to classic TD; because the step sizes were chosen to minimize the RMSE for the full experimental data, their choice inherently provided some information, which TIDBD did not receive, about the data. In a real-world application, providing this advantageous information in the form of parameters would not be possible, as the learner would be constantly faced with new, unknown data after the parameters have been set.

Despite this advantage, TIDBD and classic TD performed comparably with respect to the RMSE. This result indicates that TIDBD can act as an alternative to tuned classic TD learning, without the time- and labour-intensive setup that TD learning requires for tuning.

The choice of parameters appears to have a tremendous impact on the learning performance. Wrong parameters might result in almost no learning at all or constant overshooting. Adapting the parameters based on the incoming data should therefore result in better and more steady performance. To see whether TIDBD demonstrates more steady performance, we considered the sensitivity of the RMSE to the parameter settings for each algorithm. Our experimental data shows that classic TD was indeed strongly dependent on the learning rate: the standard deviation of the RMSE over the $264$ combinations of parameters in our full factorial design was $\sigma_{\textit{TD}\_264} = 43,734.46$. In comparison, once the best step sizes for classic TD were preselected, the standard deviation for the remaining $24$ experiments was $\sigma_{\textit{TD}\_24} = 313.42$. This value is over $100$ times smaller than the standard deviation for all $264$ experiments.

TIDBD, for which there are no learning rates to tune, attained a standard deviation of $\sigma_\textit{TIDBD} = 1,507.24$ over the $24$ Kanerva coder parameters. This value is ${\sim}30$ times smaller than $\sigma_{\textit{TD}\_264}$, but ${\sim}5$ larger than $\sigma_{\textit{TD}\_24}$. The difference in the standard deviation between classic TD with a preselected learning rate and TIDBD is most likely due to the duration of the experiment. As classic TD was initialized with optimized step sizes, it was able to perform more effectively over a short period of time.


The above comparison of classic TD and TIDBD in terms of RMSE is valuable because it helps us understand the performance of TIDBD and demonstrates its potential for attaining similar performance to classic TD without a manual tuning process for the learning rate. However, feasibility also depends on computation and memory, and there is an associated cost with using TIDBD to update the step sizes without human interaction.
For each weight in the $108$ GVFs, an additional step size was required. Given a feature representation with $30,000$ features per GVF, $3,240,000$ step sizes were required in this particular setting. Per GVF, three additional vectors of the same size as the number of features are required. In our Python implementation, each of the three additional weight vectors required $0.24$ megabytes, totalling to an additional $0.72$ megabytes.
The additional computation for updating this larger number of step sizes increased the time for updating all GVFs from $0.025$ seconds to $0.28$ seconds. However, as this corresponds to nearly four updates per second, it was still within the requirement for a prosthetic limb. 
\begin{figure} [!t]
        \centering
        \includegraphics[width = 1\textwidth]{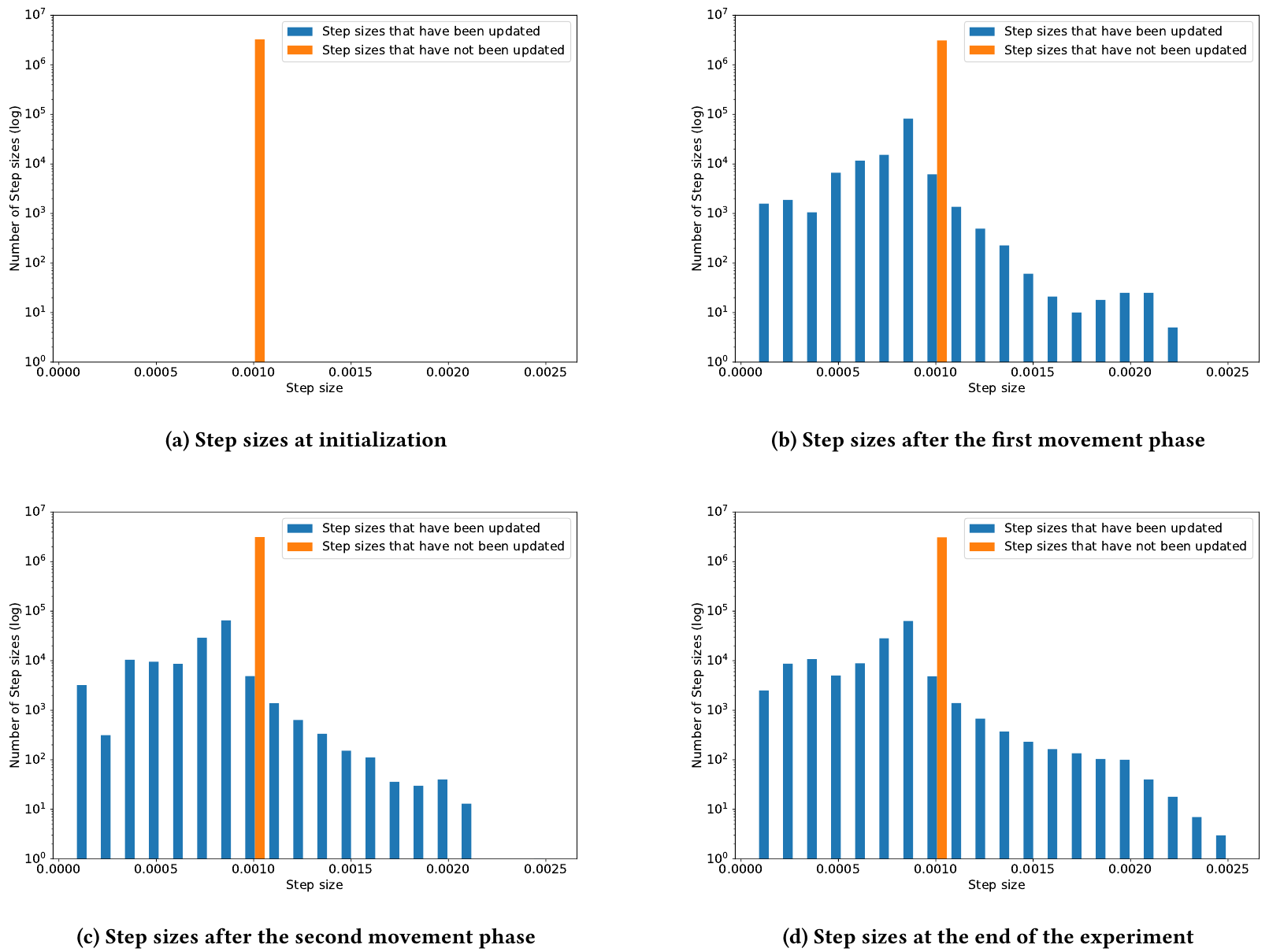}
        \caption{Step-size development over the course of the experiment. As TIDBD adapts the step sizes, this distribution will change. Subplot (A) shows the step sizes at initialization. Subplot (B) shows the step-size distribution after the first movement phase. Subplot (C) shows the step-size distribution after the second movement phase. Subplot (D) shows the step-size distribution at the end of the experiment.} 
        \label{fig:step_size_develop}
\end{figure}

The computations were performed using a Linux Mint $18.3$ OS system with an i$7$-$7700$HQ CPU with a $3.80$GHz clock rate, $6$ MB of shared L$3$ cache and $32$GB DDR$4$ RAM. With the ongoing evolution of hardware, we expect it to become possible to maintain and update even greater numbers of GVFs or to reduce the time needed for computation.

Our experimental data also offers us the opportunity to gain insight into the meta-learning process resulting from applying TIDBD. 
TIDBD assigns different step sizes to different GVFs and different features. As a result, different features contribute different amounts over time. Step sizes that are related to unimportant or noisy features will be reduced. These individual updates can be interpreted as a feature selection mechanism---TIDBD actively adapts its representation of the predictive problem, solely based on interactions with the environment \citep{kearney_tidbd_2019}.

To better understand how TIDBD changes step sizes throughout the experiment, 
Figure \ref{fig:step_size_develop} shows four snapshots of the distribution of the step sizes
. In each subplot, the orange bar shows step sizes that had not yet been updated, due to the corresponding features not being activated; the blue bars represent the step sizes that had been updated by TIDBD. Subplot (a) shows the step sizes at initialization. All of the step sizes were initialized to $0.00104$. As we would expect, Subplots (b), (c) and (d) show that the longer the experiment had run, the more the step sizes had spread out. Subplot (d) shows that the step sizes were set, by the end of the experiment, to within the range from $\num{8.0008e-05}$ to $0.00255$.

Although TIDBD actively improves its representation by adapting the step sizes, it was still sensitive to the representation that was provided, as its performance considerably varied ($\sigma_\textit{TIDBD} = 1,507.24$) with the Kanerva coder parameters---information that is not provided to the learner due to an insufficient state representation cannot be compensated for.
Within the realm of robotics, the state representation is often negatively impacted by damage to the sensors. We explore this problem in the following section by comparing the behaviour of TIDBD on data with simulated broken and stuck sensors with the behaviour demonstrated in the first experiment, shown in Figure \ref{fig:step_size_develop}.

\subsection{Stuck Sensors}
\label{stuck_sensors}
\begin{figure}[!t]
        \centering
        \includegraphics[width = 1\textwidth]{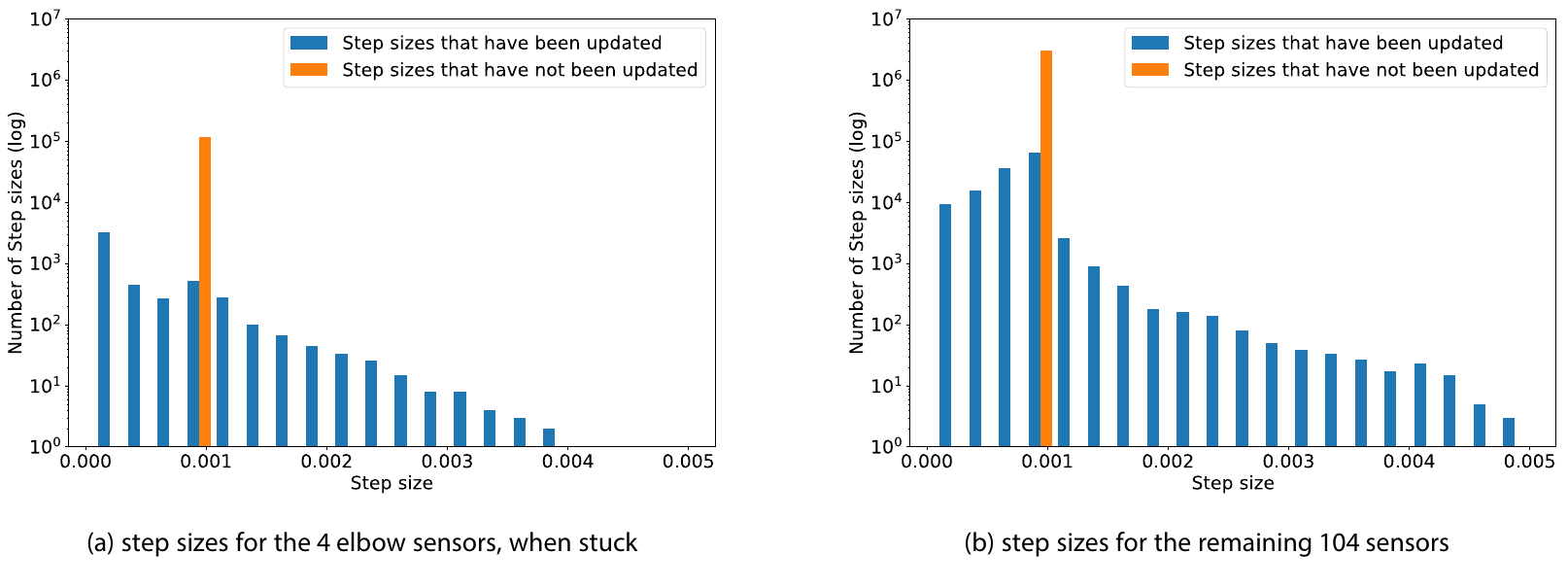}
        \caption{Step sizes distribution for the four elbow sensors (a) and the remaining 104 sensors (b), when the four elbow sensors are stuck. As described in Subsection \mbox{\ref{stuck_sensors}}, the step sizes increase noticeably compared to the original experiment. The biggest step sizes are two times as big.}
        \label{fig:step_size_stuck}
\end{figure}

For the second experiment, recall that the elbow sensors were replaced with low-variance Gaussian noise, $\mathcal{N}(1,0.5)$, to simulate them being stuck. The distribution of adapted step sizes at the end of this experiment can be found in Figure \ref{fig:step_size_stuck}.
%

In comparing Figure \ref{fig:step_size_stuck} with Figure \ref{fig:step_size_develop}, of particular note is the fact that, with simulated stuck senors, some step sizes were adapted to be much larger than any adapted during normal operation; the maximal step size when all sensors were functioning was $0.00255$, while Figure \ref{fig:step_size_stuck} shows step sizes of up to $0.005$, approximately twice as large.

The step sizes for both the predictions with stuck sensor signals as their cumulants \textit{and} for the remaining ``unaffected'' predictions increased in magnitude. 
This result may be counterintuitive at first. For a constant signal with a small amount of noise, we would expect the step sizes to decrease, as such a signal does not contain a significant amount of information. In the setting at hand, this reaction is countered by the choice of representation. As the Kanerva coder prototypes were randomly distributed in space, the small amount of noise could actually be expected to constantly lead to different prototypes being activated. At the same time, the cumulants were staying nearly constant, due to the variance being small.
This discrepancy between almost stationary cumulants and a changing representation appears to have lead to increasing step sizes, as TIDBD tried to achieve the necessary updates in fewer steps. Each feature was assigned a higher value, likely due to these updates being distributed over a wider range of features, resulting in higher step sizes. While these increasing step sizes did not necessarily improve the representation, they are clearly distinguishable from step sizes that occurred during the normal functioning of the robotic arm, thus providing important knowledge about the sensor failure.

\subsection{Broken Sensors}
\label{brocken_sensors}
The problem of broken sensors is a common one in robotics and of high interest in long-term autonomous systems. For the final experiment, the four elbow sensors were replaced with Gaussian noise, $\mathcal{N}(0,10)$, which corresponds to broken sensors that output noise. Such broken sensors do not contain meaningful information, as their output will be purely random---we therefore expect TIDBD to decrease the corresponding step sizes for these sensors.
\begin{figure}[!t]
        \centering
        \includegraphics[width = 1\textwidth]{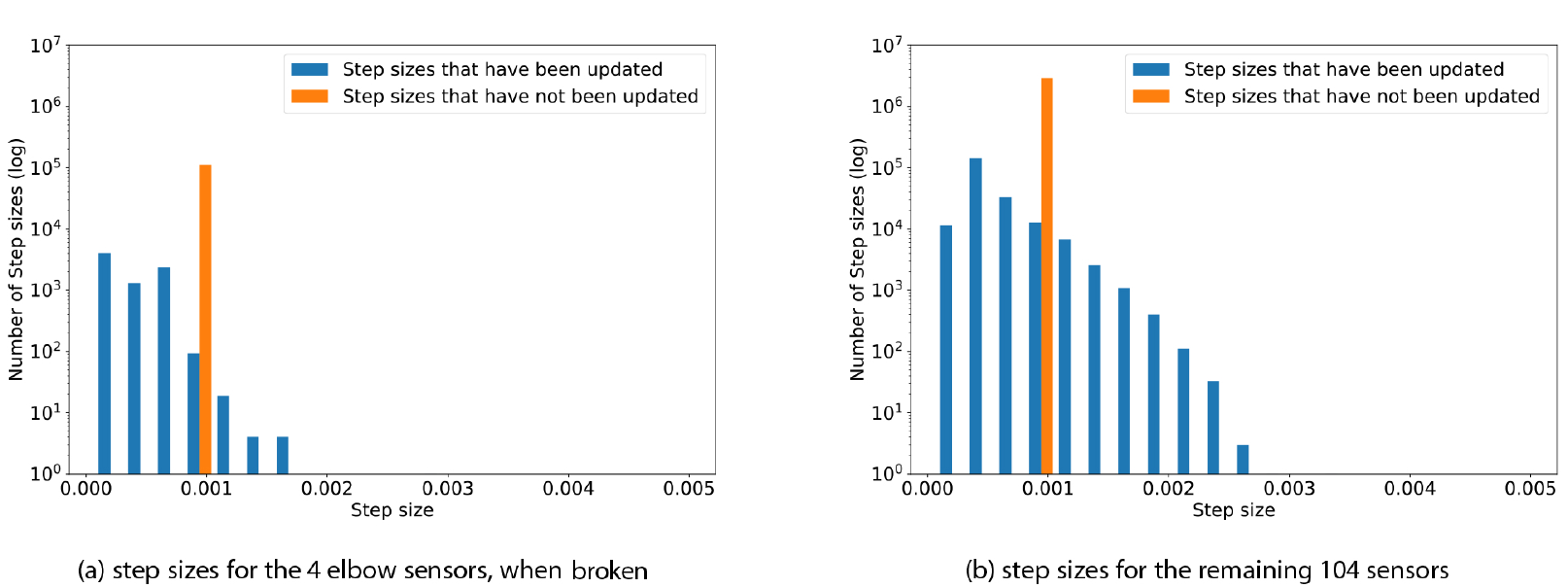}
        \caption{Step sizes distribution for the four elbow sensors (a) and the remaining 104 sensors (b), when the four elbow sensors are broken. The step sizes for the four broken sensors are noticeably reduced when compared to the experiment without broken sensors.}
        \label{fig:step_size_broken}
\end{figure}
Figure \ref{fig:step_size_broken} shows the step size distribution for the experiment with broken sensors that output high-variance noise drawn from $\mathcal{N}(0,10)$. Subplot (a) depicts the step-size distribution for the four sensors that output noise. The maximum step size is only $0.0017$. The step sizes observed during this experiment were considerably smaller than they were in the experiments where all sensors function well. The average step size for the broken sensors is $0.00037$, while the average step size for these four sensors in the experiment with functioning sensors is $0.00065$.
Subplot (b) shows the distribution for the remaining $104$ sensors. While the maximum in this experiment, with a value of $0.0028$, was almost identical to the maximum of $0.0025$ in the experiment where all sensors work well, there is a considerably distinction in the average step sizes. For the experiment with broken sensors, the average step size was $0.0006$, while it was $0.00077$ in the experiment where all of the sensors are functioning as expected. 
The RMSE for the $104$ functioning sensors, given broken elbow sensors, was calculated for both a TIDBD Horde and a classic TD Horde. The information provided by the elbow sensors was used in the feature representation $\textbf{x}(s)$, but since these sensors are broken, they only provided irrelevant, distracting information to the predictors. 
For the classic TD Horde, the RMSE for the $104$ functioning sensors increased to $1,315,850.16$. 
Step-size adaptation using TIDBD resulted in a considerably lower RMSE of $509,220.75$ in this experiment.

As expected, the step sizes corresponding to the four sensors that were replaced by noise were considerably decreased when compared to the step sizes during normal operation. Based on the interaction with these sensors, TIDBD appears to decide that it cannot learn additional information about them and to exclude them from further learning. The step sizes for the remaining $104$ sensors remained almost the same as in the normal operation of the arm. However, the distribution of step sizes in the intact sensors changed slightly as more step sizes were decreased in value---potentially to exclude features that correspond to the noisy inputs from impacting the predictions about the functioning sensor values. The RMSE for the remaining $104$ sensors supports this intuition, as it is ${\sim}2.5$ times lower ($1,315,850.16$ for classic TD vs. $509,220.75$ for TIDBD) for TIDBD than for classic TD.

\subsection{Parameter Sensitivity for TD and TIDBD}
\label{sec:parameter_sensitivity}
\begin{figure}[!ht]
    \centering
    \begin{minipage}{0.49\textwidth}
    \includegraphics[width=1.125\textwidth]{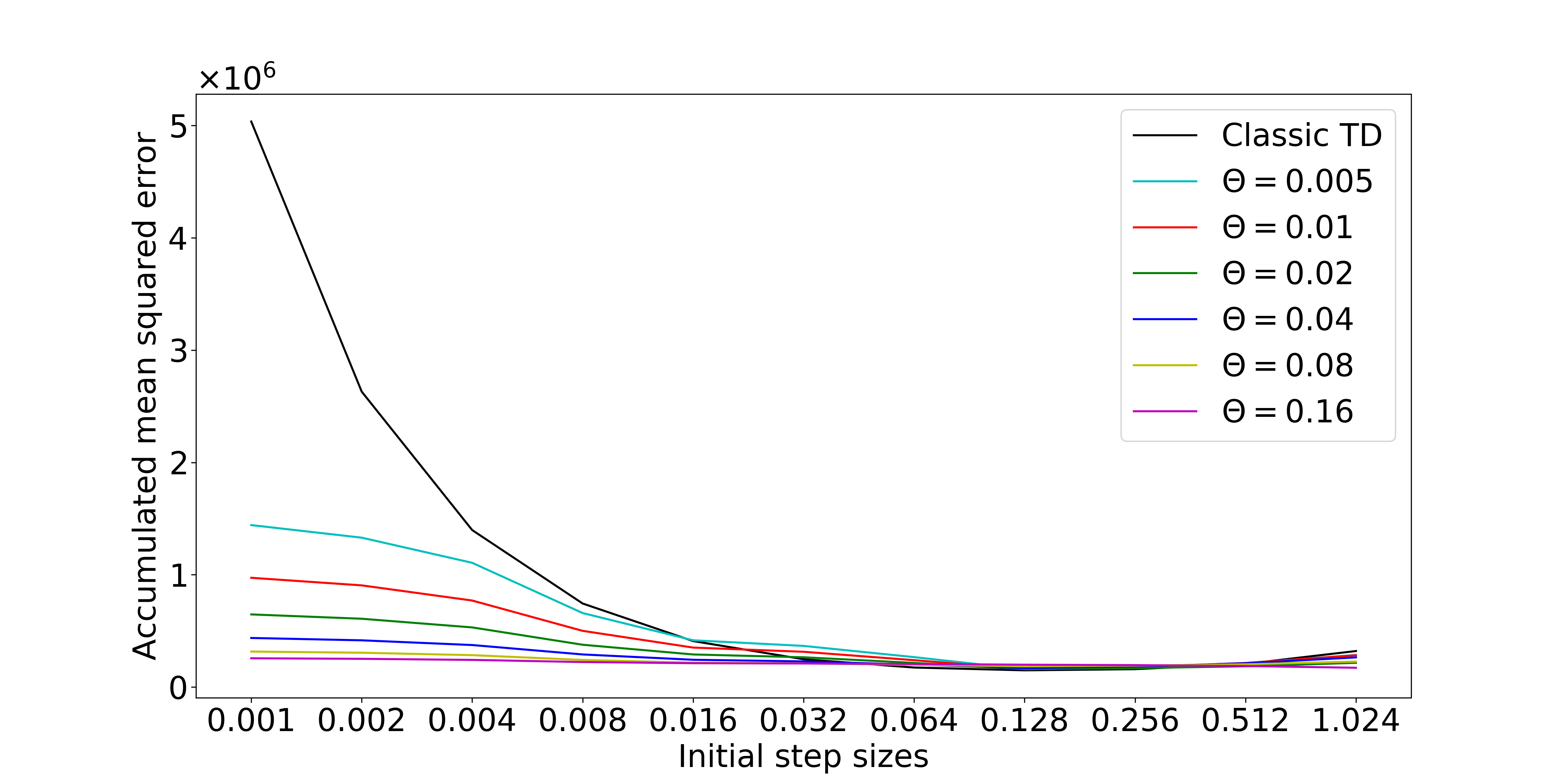}
    \centering
    \tiny (a) Accumulated RMSE over all tested initial step sizes
    \end{minipage}
    \begin{minipage}{0.49\textwidth}
    \includegraphics[width=1.125\textwidth]{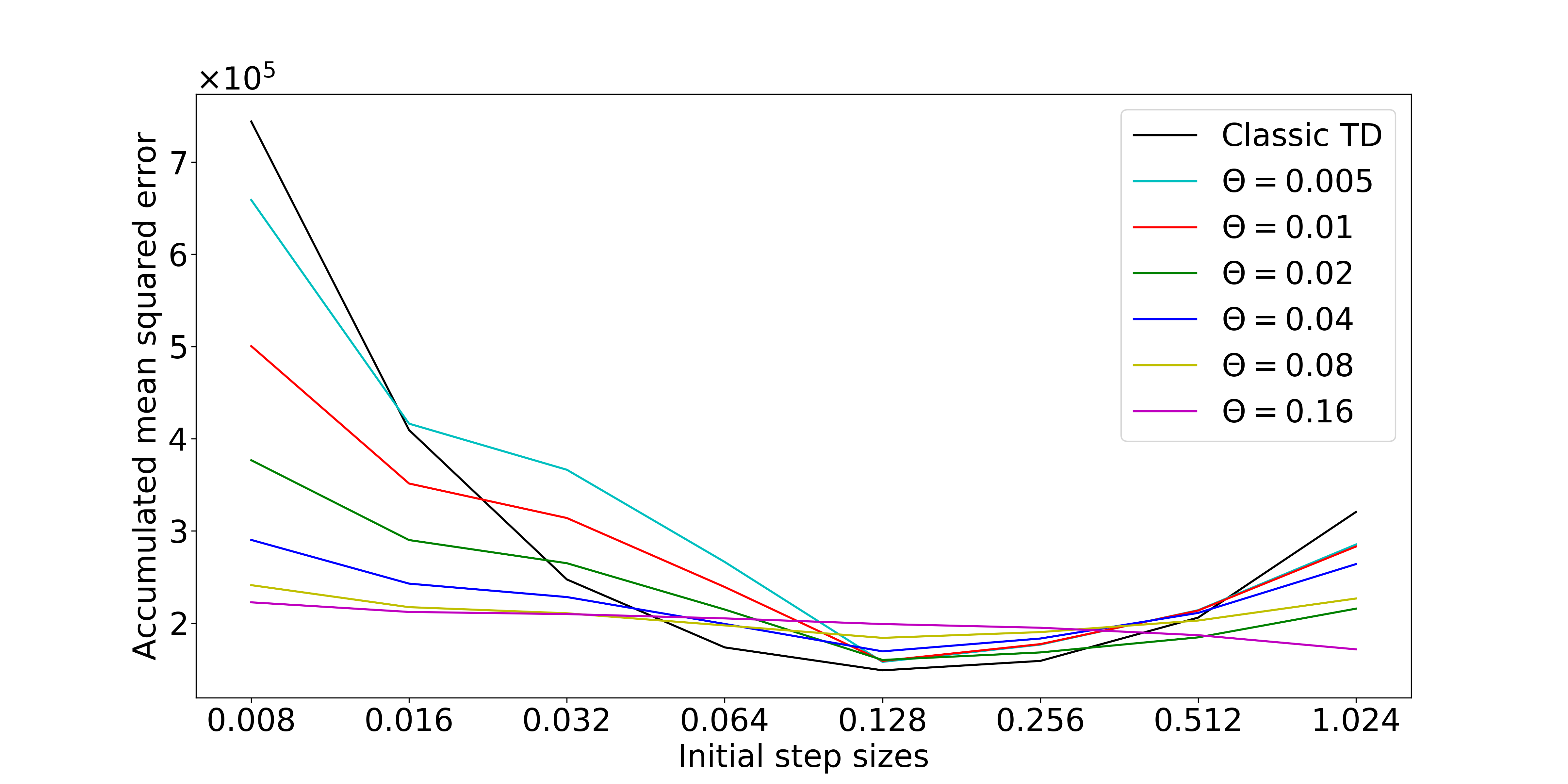}
    \centering
    \tiny (b) Zoom-in for selected initial step sizes
    \end{minipage}
    \caption{Accumulated RMSE over the experiment, depending on the initial step size. The first plot shows the overall accumulated error over the whole range of tested step sizes for TD and TIDBD with different meta step sizes $\theta$. While the performance of TD dramatically worsens for small step sizes, TIDBD exhibits more consistent and better behaviour for different meta step sizes. Subplot (b) zooms in on larger step sizes to highlight the typical bowl-shaped performance line for TD. While the error for TD is slightly smaller with carefully tuned step sizes, TIDBD shows more robust performance with respect to the initial step sizes and the meta step sizes.}
    \label{fig:step_size_sensitivity}
\end{figure}
Using TIDBD to automatically adapt parameters requires the use of meta-parameters. To investigate the sensitivity of TIDBD with respect to initial step sizes $\alpha$ and the newly introduced meta step size $\theta$, we performed sweeps over different initial values of step sizes and different meta step sizes. The results are shown in Figure \mbox{\ref{fig:step_size_sensitivity}}. Shown are the accumulated RMSEs for the whole experiment for TD and TIDBD with different values for the initial step sizes. As expected, the performance for classic TD resembles a bowl, clearly visible in the second plot. It performs poorly for very small step sizes and very large step sizes. Between these two extrema, there is a small window of good performance that traditionally has to be found by a parameter sweep.

TIDBD, however, shows much more robust performance with respect to initial parameters. While the performance is more steady over different parameters, there is still a difference for different meta-step sizes. Although this is notable, the most important takeaway from this plot is the big gap in performance between TIDBD (with any meta step size) and classic TD. The bowl shape for TIDBD is an artifact of the experiment, not of the algorithm.
The experiment has a finite amount of data which results in theoretically ideal learning rates, minimizing the error for this finite data. The closer the initial learning rates are to the ideal learning rates, the less TIDBD has to adapt the learning rates to match the ideal learning rates. This will result in a smaller error. For this reason, it makes sense that a step size adaptation algorithm with a higher meta-step size ($\theta=0.16$) will perform better on a limited data set, as it can adapt the learning rates in fewer times steps.

More importantly, the steadier performance for $\theta=0.16$ does not mean better performance when it comes to long-term error. In fact, TIDBD with lower meta step sizes demonstrates lower error for some initial step sizes. This may indicate that the meta step size $\theta=0.16$ is in fact too aggressive to converge. Therefore, the key message is the big difference between TIDBD (with any meta-step size) and classic TD. These plots confirm the results of \mbox{\cite{kearney_tidbd_2019}} and show that the meta step size and the initial step sizes can be used as default parameters within a reasonable range.

Together, the results in this paper not only support the usability of TIDBD to independently learn and update step sizes for predictions without the need of human assistance, but furthermore to independently adapt the representation that is used for a predictive knowledge approach. As TIDBD updates the step sizes based solely on interactions with the environment and is grounded in the observations that are received from said environment, it can truly function on its own---even when implemented in a long-lived application.

\section{Conclusion}
The experiments in this paper were conducted to investigate TIDBD---a step-size adaptation algorithm that assigns individual step sizes on the feature level. Four different experiments were performed on a sensor-rich robotic arm to gain further insight into the functioning of TIDBD. All four experiments utilize the data from alternating patterns of rest and movement. These four experiments result in three contributions: First, we demonstrate TIDBD to be a practical alternative to an extensive step size parameter search. Secondly, we show how TIDBD can be used to detect and characterize common sensor failures. As a third contribution, we explore TIDBD's sensitivity to its meta step size and to its initial step sizes in comparison to classic TD.

\noindent \textbf{First experiment:} We compared the predictive performance of classic TD with an extensive parameter search to the predictive performance of TIDBD. 
The additional computation required by TIDBD was still within our requirements for real-time computation and the memory used for TIDBD is negligible on modern systems. 
The results show that TIDBD and classic TD performed comparably in terms of the root mean squared error (RMSE). Although there is a set of fixed step sizes for which classic TD exhibits slightly less error on our data set, we expect TIDBD to perform better when applied in a lifelong learning setting. We expect the performance of TIDBD to exceed that of classic TD in long-term settings due to TIDBD exhibiting more improvement over time than classic TD in our experiments. These results therefore suggest that an extensive learning rate parameter search is needless.

\noindent \textbf{Second experiment:} We then explored the changes in the learning rates with several stuck sensors. The changes in the TIDBD step sizes were clearly distinguishable from changes seen during normal functioning of the arm (as explored in the first experiment), therefore providing an indicator to detect this type of sensor failure.

\noindent \textbf{Third experiment:} We replaced several sensors with high variance noise, simulating broken sensors. TIDBD decreased the step sizes corresponding to the broken sensors, which resulted in these inputs being gradually excluded from the updates---it automatically learned the unimportance of these inputs.

\noindent \textbf{Fourth experiment:} We investigate the performance in terms of accumulated RMSE for TD and TIDBD when initialized with step sizes of different magnitudes. Furthermore, the performance of TIDBD is evaluated for different meta step sizes. While the performance for TD shows a huge dependence on the initial step size value, TIDBD is more robust towards these initial step sizes and its meta step size. This shows that the usage of TIDBD is more robust with respect to its initialization, making it a viable alternative to an extensive parameter search.

These four results---the permanent updates of step sizes to accommodate non-stationarity, the distinct reaction to stuck sensors, the automatic feature selection for uninformative sensors and the robustness with respect to its initialization---are promising key features for long-term autonomous agents. They empower an agent to not only adapt its learning based on interactions with its environment, but to evaluate and improve its own perception of said environment.

Furthermore, as the step sizes contain information about the past for each feature, they can provide an important source of information to the agent itself to learn from. As has been argued prior to this work \citep{schultz2000neuronal, sherstan_introspective_2016, gunther2018predictions}, these introspective signals provide a helpful source of information to enable an agent to better understand its environment and its own functioning within its environment 

The insights presented in this paper provide deeper understanding and intuition about the effects of TIDBD, aiming to help other designers in creating agents that are capable of autonomous learning and adaptation through interaction with their environment.

\section*{Acknowledgements}
This research was undertaken, in part, thanks to funding from the Canada Research Chairs program, the Canada Foundation for Innovation, the Alberta Machine Intelligence Institute, Alberta Innovates, and the Natural Sciences and Engineering Research Council. The authors also thank Jaden Travnik for discussions on Kanerva coding and Craig Sherstan as well as Richard Sutton for suggestions and helpful discussions.

\end{document}